\documentclass{article}
\usepackage{spconf,amsmath,graphicx}
\usepackage{amssymb,amsfonts}
\usepackage{mathtools}
\usepackage{caption, subcaption}
\usepackage{paralist, tabularx}
\usepackage{siunitx}
\usepackage{booktabs}
\usepackage{xspace}
\usepackage{url}
\usepackage{microtype}

\usepackage{multicol, multirow}
\usepackage{subfiles}
\title{Embedding Compression for Teacher-to-Student Knowledge Transfer}
\name{Yiwei Ding \qquad Alexander Lerch}
  
\address{Music Informatics Group, Georgia Institute of Technology, USA}

\begin{document}
\ninept
\maketitle
\begin{abstract}
Common knowledge distillation methods require the teacher model and the student model to be trained on the same task.
However, the usage of embeddings as teachers has also been proposed for different source tasks and target tasks.
%However, embedding models\alex{is 'embedding models' a common term?} are often trained on different tasks from the target task.
Prior work that uses embeddings as teachers ignores the fact that the teacher embeddings are likely to contain irrelevant knowledge for the target task.
To address this problem, we propose to use an embedding compression module with a trainable teacher transformation to obtain a compact teacher embedding.
Results show that adding the embedding compression module improves the classification performance, especially for unsupervised teacher embeddings.
Moreover, student models trained with the guidance of embeddings show stronger generalizability.
\end{abstract}
\begin{keywords}
knowledge transfer, embedding compression, knowledge distillation
\end{keywords}

\section{Introduction}
\label{sec:intro}

The increasing model complexity of state-of-the-art deep learning approaches requires an increasing amount of training data. This progress has been enabled by the availability of large-scale datasets, as well as through progress in the development of approaches for self- and unsupervised learning, reducing the requirements for human annotations.
%In the recent decade, the availability of large-scale datasets has made it possible to train large neural networks.
%At the same time, the development of unsupervised learning has reduced the requirement for human annotations to train large models and has therefore made the training more scalable than in a fully-supervised setting.
However, there are cases where computational resources are limited, e.g., on mobile devices. Similarly, there are tasks without an abundance of training data, potentially constraining model complexity and performance. The former problem has been addressed by knowledge distillation approaches, while the latter has been addressed by transfer learning methods. %or and models cannot be arbitrarily large like mobile device applications.
%Therefore, knowledge distillation has been widely applied to compress the model.

Classical knowledge distillation requires the high-capacity teacher model to be trained on the same task or dataset as the lightweight student model \cite{hinton2015distilling, romero2015fitnets}.
There exist scenarios, however, in which the source task for teacher training is different from the target task for student training.
% However, in many scenarios, the teacher model is trained on a large dataset to make full use of its strong capacity, and the student model targets a much smaller dataset.\alex{I don't follow here --- first you say it's the same dataset, and then you say it's a large and a smaller dataset?}
% Moreover, when the teacher model is trained in an unsupervised setting to make it scalable, it becomes hard for the student model to leverage both the task-agnostic knowledge in the teacher model and the information in the target dataset.
{In this case, we can apply transfer learning before knowledge distillation by first fine-tuning the large model and then using it as a teacher model, or doing the same in reverse order.}
%Applying transfer learning might be a solution where we can first fine-tune the teacher model on the target task or dataset, and then conduct knowledge distillation \footnote{Switching the order also works but in most literature, it is applied only when the teacher is trained in a supervised setting.}.
Yet, fine-tuning a large model is a non-trivial task due to the domain shift and potential feature distortion or catastrophic forgetting, especially when there is a large dissimilarity between the source task and the target task \cite{trivedi2023closer, kumar2022fine}.
Linear probing, which refers to freezing the backbone of a model and training only the last layer can reduce these problems but might lead to suboptimal performance.
In fact, it has been shown that neither fine-tuning nor linear probing offers a one-for-all solution for transfer learning and there is no clear evidence that one outperforms the other \cite{minz2023foundation}.

\begin{figure}[!t]
 \centering
 \includegraphics[width=0.4\textwidth]{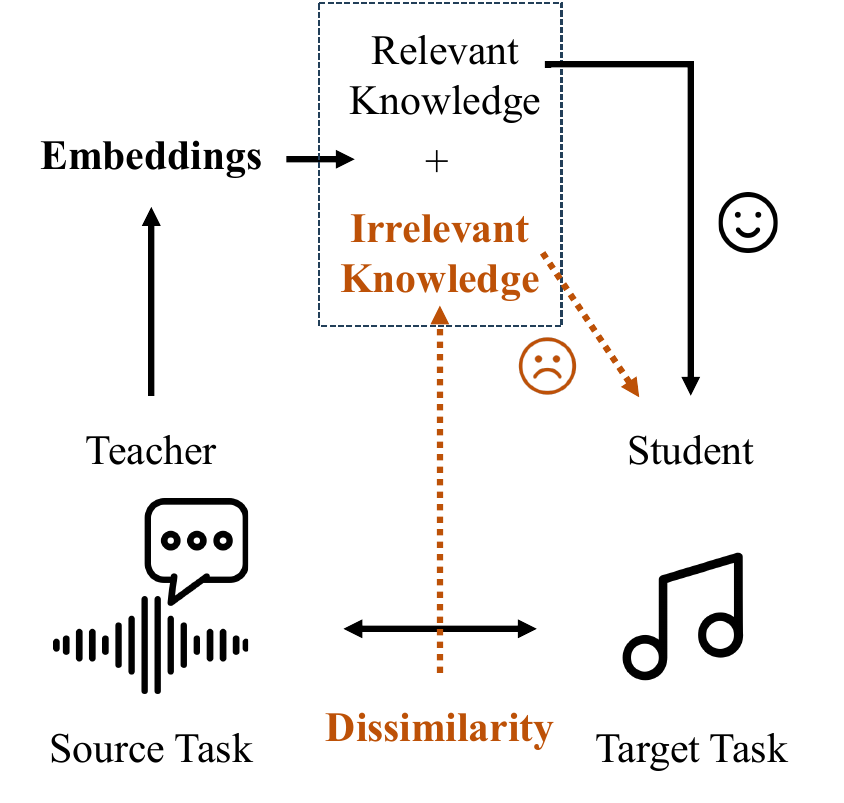}
 \caption{Illustration of irrelevant knowledge in teacher embeddings, which might make the knowledge transfer from the embeddings to the student models problematic. It is caused by the dissimilarity between the source task and the target task.
 % \alex{Not a fan of the smilies and the dotted arrows. I wonder if it would be possible to have two of these visualizations next to each other, one showing related tasks and a low amount of irrelevant information, another showing dissimilar tasks with much more irrelevant information?}
 }
 \label{fig:discrepancy}
\end{figure}

Given these challenges of adapting a model from one task to another, a recently proposed method, named Embeddings As Teachers (EAsT), aims to transfer the knowledge from the teacher model to the student model without fine-tuning or linear probing \cite{hung2022feature, ding2023audio}.
More specifically, the embeddings of large teacher models trained on the source task are used to guide the learning of the student models for the target task. This approach has been shown to improve performance on several audio and music classification tasks.
However, as is illustrated in Figure~\ref{fig:discrepancy}, 
this method does not take into account the fact that with increasing dissimilarity between source and target tasks, an increasing portion of the information in the teacher embedding may be irrelevant for the target task. This is especially the case when the teacher embeddings are trained in an unsupervised way. %there might be some irrelevant knowledge in the teacher embeddings for the target task because the embedding models are trained on different tasks from the target task and there can be a non-trivial discrepancy between the two tasks, especially when the teacher embeddings are trained in an unsupervised way and we apply them to a supervised task.
% However, this method ignores the fact that there might be some irrelevant knowledge in the teacher embeddings for the target task because the embedding models are trained on different tasks from the target task and there can be a non-trivial discrepancy between the two tasks, especially when the teacher embeddings are trained in an unsupervised way and we apply them to a supervised task.
As this irrelevant knowledge is likely to interfere negatively with the student training,
%\yiwei{This irrelevant knowledge causes the student models regularized by the unsupervised embeddings to be outperformed by the supervised ones.
%[This point actually from the results section, but is referring to the later sections in the introduction a little bit weird?]}
%Therefore, to broaden the scope of this method, 
we propose to extend the EasT approach by adding %a modified pipeline where we add 
an embedding compression module to make the embedding more compact and more relevant to the target task, which extends the usability of the method to less related tasks and unsupervised teacher embeddings. %allows the method to be used with not only supervised embeddings but unsupervised ones.

The main contribution of this paper is thus %systematic studies of
\begin{inparaitem}[]
\item the introduction of embedding compression for EasT 
%\item the effectiveness of embedding compression when using embeddings as teachers, and
%\item the generalizability of the student models trained with the method.
\end{inparaitem}
with systematic studies of both the effectiveness of this compression as well as its impact on the generalizability of the student models.

\section{Method}
\label{sec:methods}

\begin{figure*}[!t]
     \centering
     \begin{subfigure}[!t]{0.33\textwidth}
         \centering
         \includegraphics[width=0.41\textwidth]{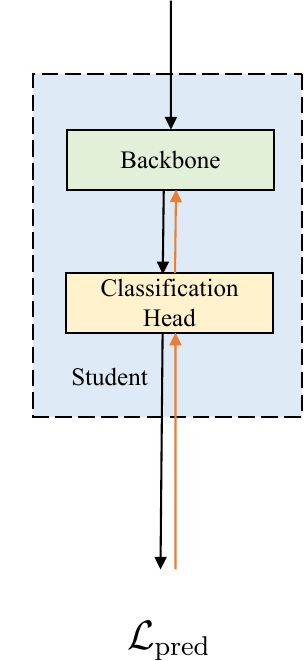}
         \caption{Training from scratch.}
         \label{fig:pipeline_scratch}
     \end{subfigure}
     \hfill
     \begin{subfigure}[!t]{0.33\textwidth}
         \centering
         \includegraphics[width=\textwidth]{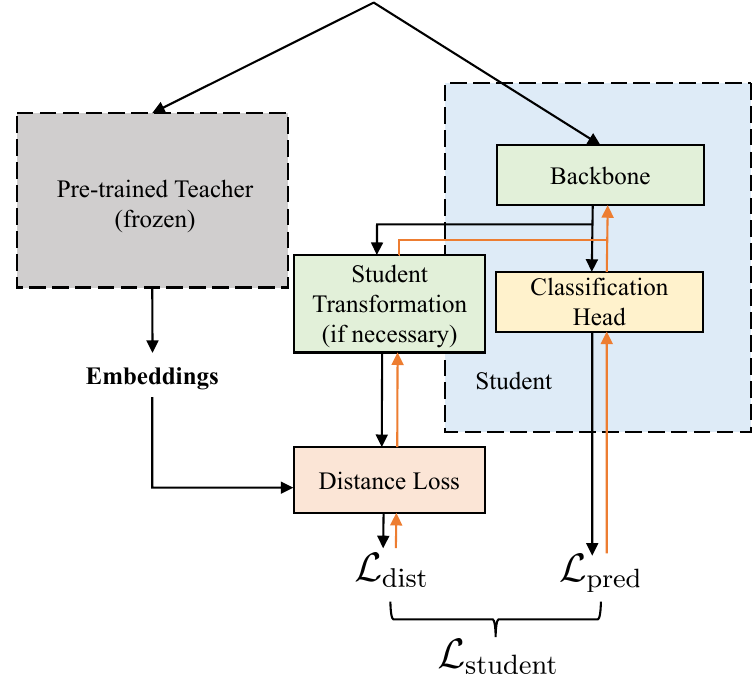}
         \caption{Training with embeddings as teachers.}
         \label{fig:pipeline_east}
     \end{subfigure}
     \hfill
     \begin{subfigure}[!t]{0.33\textwidth}
         \centering
         \includegraphics[width=\textwidth]{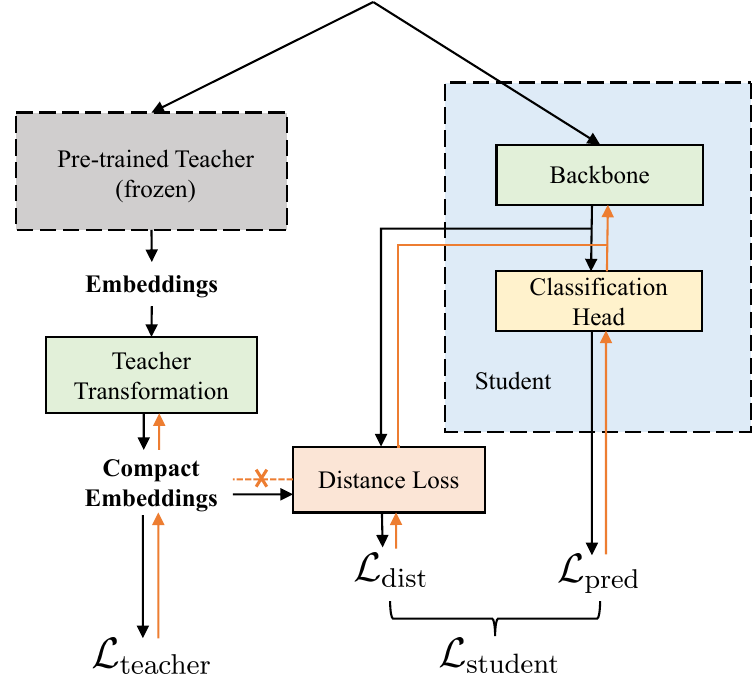}
         \caption{Training with embeddings as teachers and embedding compression.}
         \label{fig:pipeline_ceast}
     \end{subfigure}
    \caption{Different pipelines in training. The black arrows indicate the forward path and the orange arrows show the gradient flow in back propagation.}
    \label{fig:pipeline}
\end{figure*}

Figure~\ref{fig:pipeline} illustrates the pipeline of our method (\ref{fig:pipeline_ceast}) compared with training from scratch (\ref{fig:pipeline_scratch}) and directly using embeddings as teachers (\ref{fig:pipeline_east}). %, which is adopted in prior work \cite{hung2022feature, ding2023audio}.
%Since the embedding models are trained on a different task from our target task, we add the embedding compression module to make embeddings more compact, i.e., more relevant to the target task than uncompressed ones.
% The embedding compression module is trained by a separate teacher loss. \alex{Elaborate a bit here, because this teacher loss is the target task teacher loss, right? And the teacher weights remain frozen; maybe teacher loss is not the best name for it?}
More details about the embedding compression module and the distance measurement are given below.

\subsection{Embedding Compression}

To obtain a compact embedding that is more relevant to the target task than the original one, we pass the teacher embedding through a transformation to convert it into a lower-dimensional embedding.
Then the compact embedding is fed into a linear layer to obtain the teacher's prediction and to compute the loss $\mathcal{L}_\mathrm{teacher}$ on the target data.
This loss is exclusively used to update the parameters in the transformation during backpropagation; neither student nor teacher parameters are impacted by $\mathcal{L}_\mathrm{teacher}$.
% Otherwise, the teacher transformation will produce easy embeddings for students to learn, instead of those with strong prediction power for the target task.\alex{I AM NOT EXACLTY SURE WHAT THE OTHERWISE REFERS TO AND WHAT YOU MEAN BY THIS SENTENCE}
% \yiwei{Maybe adding gradient flow in the diagram could help?}

To avoid learning both the teacher transformation and the student transformation, which might lead to a collapse of the distance loss, the output dimensionality of the teacher transformation is parametrized to have the same dimensionality as the student embedding, so that a student transformation is no longer needed.

\subsection{Distance Loss}

The distance loss aims to minimize the distance between the compact embeddings and the student's output feature map so that the knowledge in the teacher embedding models can be transferred to student models.

The options investigated in this study are % can be adopted from most feature-based knowledge distillation approaches. We opt to use 
FitNet \cite{romero2015fitnets} and distance correlation \cite{szekely2007measuring}.
FitNet directly measures the Euclidean distance of two embeddings.
In the originally proposed implementation, the student embedding is first fed into a linear projection to match the dimensionality of the teacher embedding.
However, as mentioned, with embedding compression, this step can be omitted because the compact embedding already has the same dimensionality as the student.
Instead of measuring the Euclidean distance between embeddings, distance correlation measures how different the pairwise distance between samples in the two feature spaces are: if two samples are close in the teacher's embedding space, they are also supposed to be close in the student's feature space, and vice versa.
It is independent of feature dimensionalities.
% \yiwei{BUT IT DOESN'T MATTER WHETHER IT IS INDEPENDENT OF FEATURE DIMENSIONALITIES. I MENTION FITNET BECAUSE IT'S NOT THE ORIGINAL FITNET, BUT THE DISTANCE CORRELATION IS STILL THE ORIGINAl DISTANCE CORRELATION.}
More details can be found in \cite{szekely2007measuring, ding2023audio}.

\section{Experimental Setup}
\label{sec:experiments}

\begin{figure*}[!t]
     \centering
     \begin{subfigure}[!t]{0.9\textwidth}
         \centering
         \includegraphics[width=\textwidth]{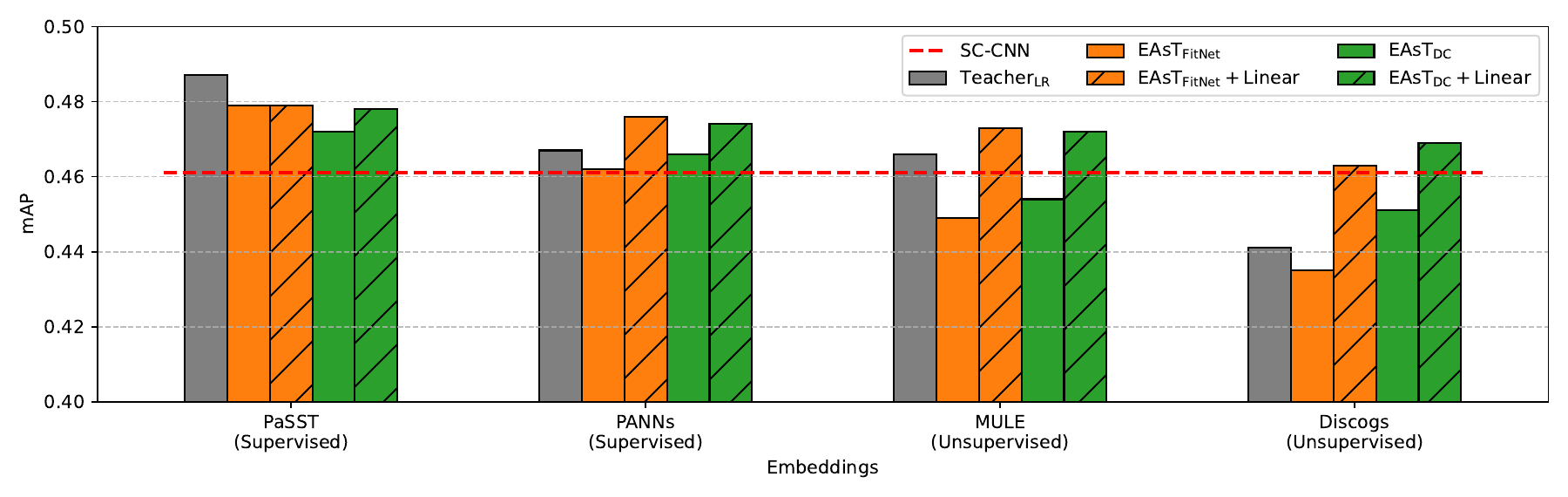}
         \caption{Results with SC-CNN.}
         \label{fig:results_shortchunk}
     \end{subfigure}
     \hfill
     \begin{subfigure}[!t]{0.9\textwidth}
         \centering
         \includegraphics[width=\textwidth]{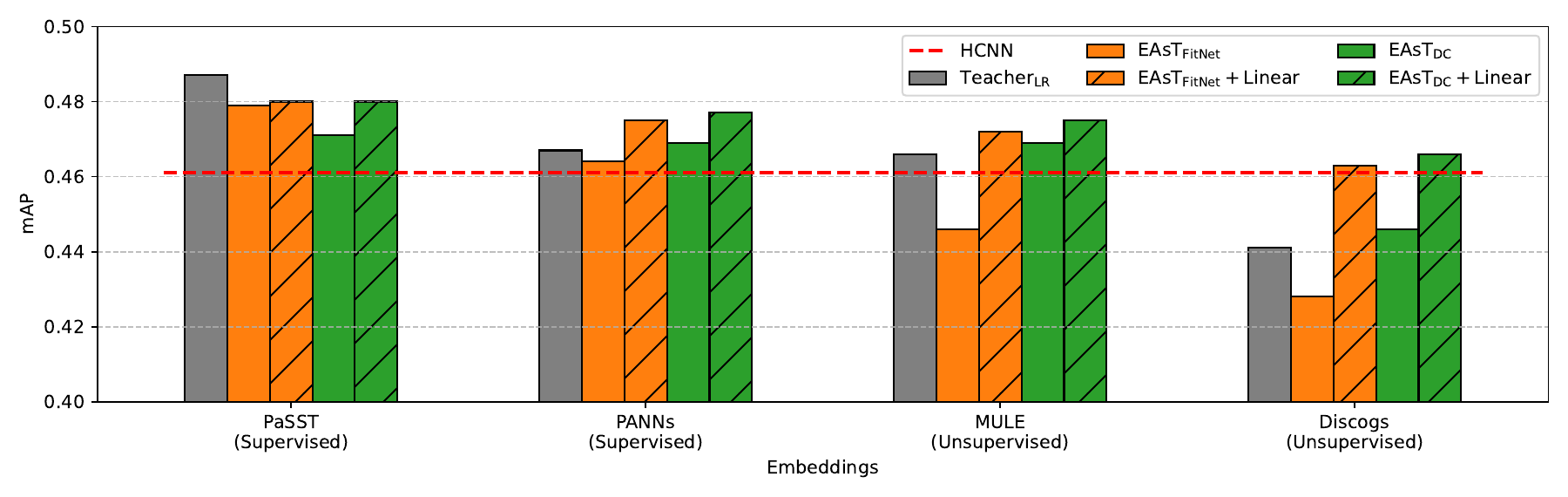}
         \caption{Results with HCNN.}
         \label{fig:results_hcnn}
     \end{subfigure}
    \caption{Results on MagnaTagATune dataset with (a) SC-CNN and (b) HCNN. Better viewed in color. The red dashed line is the baseline result. The gray bars are the results of Teacher\textsubscript{LR}. The orange bars and green bars are FitNet and distance correlation respectively. Slashed bars are those with embedding compression.}
    \label{fig:results_magnatagatune}
\end{figure*}

In this section, we describe the student models and the teacher embeddings, and then different experimental setups we use.

\subsection{Models and Embeddings}
We use Short-chunk CNN with residual connection (SC-CNN) \cite{won2020eval} and Harmonic CNN (HCNN) \cite{won2020data} as baseline models.
To the best of our knowledge, these models represent the current state-of-the-art music auto-tagging models without pre-training on extra data.

We use four different teacher embeddings for our experiments:
\begin{compactitem}
    \item \textbf{PaSST} \cite{koutini2022efficient} uses a seven-layer vision transformer that is first trained on ImageNet and then transferred to AudioSet \cite{gemmeke2017audio} for Audio Event Detection (AED). Although the structure is first proposed as a sequence-to-one model for classification tasks, it has been shown to provide powerful embeddings for short audio segments \cite{koutini2022learning}.
    \item \textbf{PANNs} \cite{kong2020panns} is a 14-layer CNN network that is trained on AudioSet for AED. Despite its simple design, it has a strong performance on AudioSet, and has shown good generalizability to several downstream tasks.
    \item \textbf{MULE} \cite{mccallum2022supervised} builds on the contrastive learning framework SimCLR \cite{chen2020simple} and its audio domain application COLA \cite{saeed2021contrastive} to learn a music representation. The positive samples are created using different segments from the same audio track following COLA and thus do not require extra data augmentation as in SimCLR.
    \item  \textbf{DisCogs} \cite{alonso2022music} learns a music representation by using editorial data like artists. Specifically, two samples with overlapping artist data are considered to be a positive pair and otherwise, they become a negative pair. Its integration of metadata into contrastive learning is an interesting multimodal contrastive learning approach. 
\end{compactitem}

\subsection{Comparison with Baseline}
We first evaluate the effectiveness of our method on the music auto-tagging task with the MagnaTagATune dataset \cite{law2009evaluation}, which has 25,860 audio clips of approximately \qty{29.1}{s} in length.
We evaluate the methods in terms of mean Average Precision (mAP).
To show the broader scope of our method, we also evaluate the proposed method on sound event classification and music genre classification.
For sound event classification, we use the ESC50 dataset \cite{piczak2015dataset}, which contains 2000 audio snippets of \qty{5}{s} in fifty balanced categories.
For music genre classification, we use the FMA-small dataset \cite{fma_dataset}, which has 8000 audio tracks of \qty{30}{seconds} in 8 balanced genres.
On both ESC50 and FMA-small, we report the classification accuracy.

For a complete comparison of different approaches, we include the results of the following systems on the MagnaTagATune dataset:
\begin{compactitem}
    \item \textbf{Teacher\textsubscript{LR}} is logistic regression with the teacher embeddings.
    \item \textbf{Baseline (SC-CNN or HCNN)} is the student model trained from scratch.
    \item \textbf{EAsT\textsubscript{FitNet}} is the student model trained with FitNet distillation.
    \item \textbf{EAsT\textsubscript{FitNet} + Linear} is EAsT\textsubscript{FitNet} with the proposed embedding compression module. The teacher transformation is a linear projection. As stated in Section~\ref{sec:methods}, we remove the linear projection for the student because the compressed embedding is already the same dimensionality as the student's feature.
    \item \textbf{EAsT\textsubscript{DC}} is the student model trained with distillation loss computed with distance correlation.
    \item \textbf{EAsT\textsubscript{DC} + Linear} is EAsT\textsubscript{DC} with the proposed embedding compression module.
\end{compactitem}

For the ESC50 dataset and the FMA dataset, we use SC-CNN as the baseline student model, PaSST and MULE as teacher embeddings, and compare different EAsT methods.

We use a linear projection as a teacher transformation as we find that using a more complex transformation leads to overfitting of the teacher transformation with our target task datasets and therefore deteriorates the performance.

\subsection{Generalizability}
We test the generalizability of the student models by evaluating them on a different dataset without extra training or fine-tuning, which is close to real-world scenarios where the test data is not only inaccessible but can also be very different from the training data.
Because half of the labels in the MagnaTagATune dataset are instrument labels, we evaluate our models on the OpenMIC dataset \cite{humphrey2018openmic}, which aims to identify the musical instruments in the audio clips.
We compute the mAP (averaged among all instrument classes) for the nine overlapping instrument labels between OpenMIC and MagnaTagATune.

\section{Results and Discussions}
\label{sec:results}

In this section, we present our results with some discussions.

\subsection{Comparison with baseline}
\label{subsec:results_baseline}
Figure~\ref{fig:results_magnatagatune} shows the results on the MagnaTagATune dataset.
Comparing results with or without embedding compression, we can observe that in all cases the embedding compression module can improve the performance, except for the FitNet + PaSST combination where the result stays the same.
Comparing the improvement between supervised embeddings and unsupervised embeddings, we notice that embedding compression leads to a larger improvement when it is applied to unsupervised embeddings.
Moreover, we can see that without embedding compression, the EAsT method tends to deteriorate the performance compared with the baseline, but adding the embedding compression enables the student model to outperform the baseline.

These observations are consistent with our assumption that the irrelevant knowledge in the teacher embeddings negatively impacts the knowledge transfer.
The extent of irrelevance depends on the domain shift from the source task to the target task, which is greater in the case of unsupervised embeddings than supervised ones.
% By adding the knowledge compression module, we successfully apply the method of using embeddings as teachers to unsupervised embeddings.\alex{MOVE TO CONCLUSION}

While the transformation of the teacher embeddings shows some parallels to fine-tuning the teacher model, no parameters of the teacher model are changed. Therefore, the feature distortion is minimized and the risk of overfitting reduced. In addition, a considerable amount of computational resources can be saved during the training.

\subsection{Generalizability}

\begin{table}
    \centering
    %\begin{tabular}{l c  c  c}
     \begin{tabular*}{\columnwidth}{l @{\extracolsep{\fill}} c  c  c}
        \hline \hline
        & None & PaSST & MULE \\
        \hline
        SC-CNN
                & .863 & -- & --\\
        \hline
        EAsT\textsubscript{FitNet}
                & -- & .904          & .890\\
        EAsT\textsubscript{FitNet} + Linear
                & -- & .904          & .887\\
        EAsT\textsubscript{DC}
                & -- & .892          & .874\\
        EAsT\textsubscript{DC} + Linear
                & -- & .901          & .887\\
        \hline \hline
    \end{tabular*}
   \caption{Generalizability test. All models are trained on MagnaTagATune and tested on the overlapping labels in OpenMIC. SC-CNN trained from scratch on MagnaTagATune serves as the baseline model.}
   \label{tab:res_generlizability}
\end{table}

The results of the generalizability test are listed in Table~\ref{tab:res_generlizability}. Note that these models are not trained on the target dataset.

Comparing the EAsT methods with the baseline SC-CNN, we can see that the models trained with the guidance of embeddings show improved performance. This is true for the models with or without embedding compression. These results suggest that adding the knowledge of embeddings during training improves the generalizability of student models. The teacher models, trained on large-scale datasets and thus having better generalizability, can transfer this general knowledge to the students.

In the case of DC, embedding compression tends to slightly improve the results, and in the case of FitNet, the performance stays the same or has a slight decay.
However, the differences are relatively small compared to the improvement over the baseline, which means that while adding the embedding compression module might lead to a bias toward the target task, it does not have a significant negative impact on the student model's generalizability.

\subsection{Complexity}

Table~\ref{tab:results_complexity} lists the number of parameters and the rough inference speed measurements of the models we use.

\begin{table}
    \begin{center}
    \begin{tabular*}{\columnwidth}{l @{\extracolsep{\fill}}cc}
    %\begin{tabular}{c|cc}
    \hline
    \hline
    Model                   & Parameters (M)    & Iteration / s\\
    \hline
    PaSST                   & 86.1              & 18.7\\
    PANNs                   & 79.7              & 70.6\\
    MULE                    & 62.4              & 53.5\\
    DisCogs                 & 4.0               & 101.0\\
    \hline
    HCNN                    & 3.6               & 164.2\\
    SC-CNN                  & 9.1               & 196.4\\
    \hline
    \hline
    
    \end{tabular*}
    \end{center}
    \caption{Comparison of the model complexity.}
    \label{tab:results_complexity}
        %\vspace{-1mm}

\end{table}

We can see that the student models we use have much fewer parameters and are faster to run compared to the teacher embedding models (except for DisCogs, which is based on EfficientNet and has fewer parameters but also leads to suboptimal performance).

\subsection{Results on other datasets}

We report the performance in terms of classification accuracy on the ESC50 dataset and the FMA-small dataset in Table~\ref{tab:res_others}.
The experimental setup is the same as in Sect.~\ref{subsec:results_baseline}, i.e., the student models have been trained on the target dataset.

\begin{table}
    %\begin{subtable}{0.45\textwidth}
        \centering
        %\begin{tabular}{cccc}
        \begin{tabular*}{\columnwidth}{l @{\extracolsep{\fill}} c  c  c}
        \hline \hline
        \textbf{ESC50}   & None  & PaSST & MULE \\
        \hline
        SC-CNN
                & .732  & --     & --\\
        \hline
        EAsT\textsubscript{FitNet}
                & --     & .754  & .749\\
        EAsT\textsubscript{FitNet} + Linear
                & --     & .747  & .758\\
        EAsT\textsubscript{DC}
                & --     & .753  & .728\\
        EAsT\textsubscript{DC} + Linear
                & --     & .753  & .746\\
        \hline \hline
        \textbf{FMA-small}   & None  & PaSST & MULE \\
        \hline
        SC-CNN
                & .516  & --     & --\\
        \hline
        EAsT\textsubscript{FitNet}
                & --     & .512  & .526\\
        EAsT\textsubscript{FitNet} + Linear
                & --     & .527  & .527\\
        EAsT\textsubscript{DC}
                & --     & .505  & .494\\
        EAsT\textsubscript{DC} + Linear
                & --     & .545  & .531\\
        \hline \hline
        \end{tabular*}
        %\caption{Results on the FMA-small dataset.}
        %\label{tab:res_fmas}
     %\end{subtable}
     \caption{Results on other datasets.}
     \label{tab:res_others}
\end{table}

On the ESC50 dataset, we observe a similar trend as in Fig.~\ref{fig:results_magnatagatune} that when using the unsupervised MULE embedding, embedding compression can improve the performance.
However, in the case of supervised PaSST embedding, adding the embedding compression module shows no benefits, as the performance either deteriorates or stays the same.
A possible reason is that the dissimilarity between the source task (AED) that PaSST is trained on and the target task (sound event classification) is smaller than that between AED and music auto-tagging, therefore the embedding compression is not able to reduce much irrelevant information.
On the FMA-small dataset, we find that the embedding compression module can improve the results with both embeddings, which supports our assumption that embedding compression is more effective especially when there is a greater dissimilarity between the source task and the target task.

% \section{Related work}
% \label{sec:related_work}
% \subfile{sections/related_work}

\section{Conclusion and future work}
\label{sec:conclusion}

In this paper, we propose a novel embedding compression module for transferring knowledge by using embeddings as teachers.
This approach considers the irrelevant knowledge in the teacher embeddings caused by the dissimilarity between source tasks and target tasks and yields a performance improvement with unsupervised embeddings. Finally, we show that student models trained with the proposed method have better generalizability properties without any extra training.
In the field of audio and music deep learning, training an embedding model and applying the embeddings to downstream tasks is more popular than training different models for different tasks, mainly due to the scarcity of training data in many tasks.
Therefore, the proposed method is more suitable for this field than classical knowledge distillation.
In addition, as unsupervised learning methods are gaining increasing attention due to their better scalability, we believe the proposed method has a broad scope of application not limited to audio tasks.

% \alex{Some of your references are incomplete, please double check.}

\vfill\pagebreak

\bibliographystyle{IEEEbib}
\bibliography{refs}

\begin{thebibliography}{10}

\bibitem{hinton2015distilling}
Geoffrey Hinton, Oriol Vinyals, and Jeff Dean,
\newblock ``Distilling the knowledge in a neural network,''
\newblock {\em arXiv:1503.02531}, 2015.

\bibitem{romero2015fitnets}
Adriana Romero, Nicolas Ballas, Samira~Ebrahimi Kahou, Antoine Chassang, Carlo
  Gatta, and Yoshua Bengio,
\newblock ``{FitNets}: Hints for thin deep nets,''
\newblock in {\em Proceedings of the International Conference on Learning
  Representations (ICLR)}, 2015.

\bibitem{trivedi2023closer}
Puja Trivedi, Danai Koutra, and Jayaraman~J Thiagarajan,
\newblock ``A closer look at model adaptation using feature distortion and
  simplicity bias,''
\newblock in {\em Proceedings of the International Conference on Learning
  Representations (ICLR)}, 2023.

\bibitem{kumar2022fine}
Ananya Kumar, Aditi Raghunathan, Robbie Jones, Tengyu Ma, and Percy Liang,
\newblock ``Fine-tuning can distort pretrained features and underperform
  out-of-distribution,''
\newblock in {\em Proceedings of the International Conference on Learning
  Representations (ICLR)}, 2022.

\bibitem{minz2023foundation}
Minz Won, Yun-Ning Hung, and Duc Le,
\newblock ``A foundation model for music informatics,''
\newblock in {\em Proceedings of the International Conference on Acoustics,
  Speech and Signal Processing (ICASSP)}, 2024.

\bibitem{hung2022feature}
Yun-Ning Hung and Alexander Lerch,
\newblock ``Feature-informed embedding space regularization for audio
  classification,''
\newblock in {\em Proceedings of the European Signal Processing Conference
  (EUSIPCO)}, 2022.

\bibitem{ding2023audio}
Yiwei Ding and Alexander Lerch,
\newblock ``Audio embeddings as teachers for music classification,''
\newblock in {\em Proceedings of the International Society for Music
  Information Retrieval Conference (ISMIR)}, 2023.

\bibitem{szekely2007measuring}
G{\'a}bor~J Sz{\'e}kely, Maria~L Rizzo, and Nail~K Bakirov,
\newblock ``Measuring and testing dependence by correlation of distances,''
\newblock {\em The Annals of Statistics}, vol. 35, no. 6, pp. 2769--2794, 2007.

\bibitem{won2020eval}
Minz Won, Andres Ferraro, Dmitry Bogdanov, and Xavier Serra,
\newblock ``Evaluation of {CNN}-based automatic music tagging models,''
\newblock in {\em Proceedings of the Sound and Music Computing (SMC)}, 2020.

\bibitem{won2020data}
Minz Won, Sanghyuk Chun, Oriol Nieto, and Xavier Serrc,
\newblock ``Data-driven harmonic filters for audio representation learning,''
\newblock in {\em Proceedings of the International Conference on Acoustics,
  Speech and Signal Processing (ICASSP)}, 2020.

\bibitem{koutini2022efficient}
Khaled Koutini, Jan Schl{\"u}ter, Hamid Eghbal-zadeh, and Gerhard Widmer,
\newblock ``Efficient training of audio transformers with patchout,''
\newblock in {\em Proceedings of INTERSPEECH}, 2022.

\bibitem{gemmeke2017audio}
Jort~F Gemmeke, Daniel~PW Ellis, Dylan Freedman, Aren Jansen, Wade Lawrence,
  R~Channing Moore, Manoj Plakal, and Marvin Ritter,
\newblock ``Audio set: An ontology and human-labeled dataset for audio
  events,''
\newblock in {\em Proceedings of the International Conference on Acoustics,
  Speech and Signal Processing (ICASSP)}, 2017.

\bibitem{koutini2022learning}
Khaled Koutini, Shahed Masoudian, Florian Schmid, Hamid Eghbal-zadeh, Jan
  Schl\"{u}ter, and Gerhard Widmer,
\newblock ``Learning general audio representations with large-scale training of
  patchout audio transformers,''
\newblock in {\em HEAR: Holistic Evaluation of Audio Representations (NeurIPS
  2021 Competition)}, 2022.

\bibitem{kong2020panns}
Qiuqiang Kong, Yin Cao, Turab Iqbal, Yuxuan Wang, Wenwu Wang, and Mark~D
  Plumbley,
\newblock ``Panns: Large-scale pretrained audio neural networks for audio
  pattern recognition,''
\newblock {\em IEEE/ACM Transactions on Audio, Speech, and Language
  Processing}, vol. 28, pp. 2880--2894, 2020.

\bibitem{mccallum2022supervised}
Matthew~C McCallum, Filip Korzeniowski, Sergio Oramas, Fabien Gouyon, and
  Andreas~F Ehmann,
\newblock ``Supervised and unsupervised learning of audio representations for
  music understanding,''
\newblock in {\em Proceedings of the International Society for Music
  Information Retrieval Conference (ISMIR)}, 2022.

\bibitem{chen2020simple}
Ting Chen, Simon Kornblith, Mohammad Norouzi, and Geoffrey Hinton,
\newblock ``A simple framework for contrastive learning of visual
  representations,''
\newblock in {\em Proceedings of the International Conference on Machine
  Learning (ICLR)}, 2020.

\bibitem{saeed2021contrastive}
Aaqib Saeed, David Grangier, and Neil Zeghidour,
\newblock ``Contrastive learning of general-purpose audio representations,''
\newblock in {\em Proceedings of the International Conference on Acoustics,
  Speech and Signal Processing (ICASSP)}, 2021.

\bibitem{alonso2022music}
Pablo Alonso-Jim{\'e}nez, Xavier Serra, and Dmitry Bogdanov,
\newblock ``Music representation learning based on editorial metadata from
  discogs,''
\newblock in {\em Proceedings of the International Society for Music
  Information Retrieval Conference (ISMIR)}, 2022.

\bibitem{law2009evaluation}
Edith Law, Kris West, Michael~I Mandel, Mert Bay, and J~Stephen Downie,
\newblock ``Evaluation of algorithms using games: The case of music tagging,''
\newblock in {\em Proceedings of the International Society for Music
  Information Retrieval Conference (ISMIR)}, 2009.

\bibitem{piczak2015dataset}
Karol~J. Piczak,
\newblock ``{ESC}: {Dataset} for {Environmental Sound Classification},''
\newblock in {\em Proceedings of the Annual ACM Conference on Multimedia
  (ACMMM)}, 2017.

\bibitem{fma_dataset}
Micha\"el Defferrard, Kirell Benzi, Pierre Vandergheynst, and Xavier Bresson,
\newblock ``{FMA}: A dataset for music analysis,''
\newblock in {\em Proceedings of the International Society for Music
  Information Retrieval Conference (ISMIR)}, 2017.

\bibitem{humphrey2018openmic}
Eric Humphrey, Simon Durand, and Brian McFee,
\newblock ``{OpenMIC}-2018: An open data-set for multiple instrument
  recognition.,''
\newblock in {\em Proceedings of the International Society for Music
  Information Retrieval Conference (ISMIR)}, 2018.

\end{thebibliography}

\end{document}